# Can Artificial Intelligence Embody Moral Values?

Torben Swoboda (Vlerick Business School, KU Leuven) & Lode Lauwaert (KU Leuven)

Corresponding author: torben.swoboda@kuleuven.be

## Abstract

The neutrality thesis holds that technology cannot be laden with values – it is inherently value-neutral. This long-standing view has faced critiques, but much of the argumentation against neutrality has focused on traditional, non-smart technologies like bridges and razors. In contrast, AI is a smart technology increasingly used in high-stakes domains like healthcare, finance, and policing, where its decisions can cause moral harm. In this paper, we argue that artificial intelligence (AI), particularly artificial agents that autonomously make decisions to pursue their goals, challenge the neutrality thesis. Our central claim is that the computational models underlying artificial agents can integrate representations of moral values such as fairness, honesty and avoiding harm. We provide a conceptual framework discussing the neutrality thesis, values, and AI. Moreover, we examine two approaches to designing computational models of morality, artificial conscience and ethical prompting, and present empirical evidence from text-based game environments that artificial agents with such models exhibit more ethical behavior compared to agents without these models. The findings support that AI can embody moral values, which contradicts the claim that all technologies are necessarily value-neutral.

## 1. Introduction

One of the most well-known theses about technology is that it is neutral, also known as the neutrality thesis. The thesis has been discussed by many academics (Floridi & Sanders, 2004; Heyndels, 2023; Klenk, 2021; van de Poel and Kroes, 2014; Tollon, 2021; Pitt, 2014). Philosopher Joseph Pitt's 2014 paper titled '"Guns don't kill, people kill"; values in and/or around technologies' is widely regarded as one of the key defenses of the neutrality thesis.[2] One of Pitt's arguments is that it is impossible to perceive the alleged ladenness of an artifact. One can only see pure matter, and thus, technology is neutral. In contrast, one could say that technology *may* be neutral but it is not the case that technology is necessarily neutral.

Yet, delving into the literature on the question whether or not technology is neutral, one can see that the argumentation against the neutrality thesis is (most often) based upon technologies that are *not* equipped with artificial intelligence (AI). Examples include prehistoric inventions like a bow-and-arrow or spear, Robert Moses' 'racist' bridges as described by Langdon Winner (2017),[1] and the razors from the 1980s from the Dutch company Philips that were seen as sexist, since there were two types: one that could be repaired (for men), one that could not be repaired (for women). There are far fewer papers that focus on smart technologies within the context of the discussion of the neutrality thesis (see e.g. Gabriel and Ghazavi, 2022; van de Poel, 2020).

From one point of view, it is surprising that there's not that much academic literature about the question whether or not AI systems are neutral. Bias and discrimination are two hot topics within the realm of AI ethics, and with, for example, a leading study such as *Weapons of Math Destruction* by Cathy O'Neil (2017), scholars have become familiar with examples of intelligent systems that treat people of color differently than white people, and thus that are not neutral. And yet, that is not the end of the debate. After all, if today it is concluded that some AI is not neutral, the technology is usually said to be not neutral in an *undesirable* way, namely because the system produces sexist or racist output. Little or no attention has been paid to the ladenness of AI in a desirable way. Furthermore, the debate is also mostly about undermining what we call 'a broad interpretation of the neutrality thesis', which holds that technology is not laden with a wide range of cultural products: stereotypes, ideologies, norms, perspectives, etc. The question is generally not so much about 'a narrow interpretation of the neutrality thesis', which is only about *moral values*, and which holds that AI systems are independent from moral values, such as non-maleficence, fairness, or honesty.[2] In other words, up until now, not that much has been said about the specific question about whether AI in a desirable sense is laden with a moral value.

Focusing on this last question, at first glance it seems that it should be answered in the affirmative. For instance, privacy is a moral value and we could develop an AI system in a way that is privacy-proof, in the sense that the data used to train the system are given with consent. Consequently, one can conclude that AI is not necessarily neutral in a desirable sense.

---

[1] There has been criticism of the case described by Winner, and more about whether or not the case is truthful (Joerges, 1999; Woolgar and Cooper, 1999). However, even if not all is correct, one could consider the example of Moses' bridges as a thought experiment that indicates the problem with the neutrality thesis.

[2] Thus, our paper will not deal with non-moral values such as, for example, aesthetic values (beauty) or epistemic values (truth).

However, value-ladenness here has to do primarily with the data the system is trained with, not with a specific ability that some AI systems have, namely, to make decisions. In other words, so far, the question has not so much been on the moral ladenness of the learned AI model that produces the output. This makes clear that at least one fundamental question has been understudied in the field of AI ethics to date. Several other questions are already widely discussed, such as, to name just a few, whether it is possible to develop machine consciousness (Schneider, 2019), whether machines can think (Searle, 1980), or whether it makes sense to attribute rights to machine learning systems (Gunkel, 2018). The question that we now want to push to the forefront is the following: how (if at all) can AI be laden with a moral value in a desirable sense?

This question is relevant for at least two reasons. The first justification for examining whether algorithms can be laden with a moral value is pragmatic. For example, the limited functionality of tenant screening tools may associate arrest records with wrong individuals and facial recognition errors can lead to wrongful arrests (Raji et al., 2022). Language models, used in chatbots, produce toxic language, including racist and sexist statements (Gehman et al., 2020). Generative AI is also used to create deepfakes, i.e. synthetic media like images, audio, and video that appears highly realistic. Celebrities, like Taylor Swift, have become the target of sexual deepfakes and scammers utilize voice cloning to trick their victims more effectively (Hsu, 2024). These examples illustrate the morally harmful impact AI systems can have. One possible response to prohibit such outcomes, which is one of the central topics within the field of value alignment and that has also been discussed by Nick Bostrom in *Superintelligence* (2014), is to embed moral values into the technology (Anwar et al., 2024, pp. 79f). After all, a value-laden system carries the promise to produce output that is in line with moral values such as fairness and honesty. However, if proponents of the neutrality thesis were correct then value alignment would be conceptually impossible and research efforts in this direction futile. Thus, it is important to establish whether it is possible to design AI systems that adhere to moral values.

The second reason is to critically evaluate Pitt's argumentation by relating it to AI. More precisely, Pitt argues that values are inherently linked to decision-making and pursuing goals. People can make decisions, and since decision-making is linked to moral values, people can embody such values. But technologies like razors or sea dykes are not capable of making decisions and, therefore, they cannot be imbued with values. However, AI engages in decision-making, which undermines Pitt's argument. Therefore, it seems that AI is a special kind of technology as it possesses certain capabilities that make it more prone to be laden with values compared to other kinds of technology.

The main aim of this paper is to delve into the question whether an AI system can be laden with a moral value in a desirable way. Before we can present our argument, some preliminary conceptual work is needed. For example, we need to explain what we mean by artificial agents and why we are focusing on this type of AI system. In addition, we should clarify some point regarding the neutrality thesis and also address the functionalist conception of values on which our argument relies. This is what we will do in the second section, called 'conceptual clarifications'. The third section is argumentative in nature and also the central part of the paper. There we show that it is in principle possible that AI's algorithms and computational models are not neutral, and that has to do with an AI system's ability to choose between options. Our conclusion, thus, will be that the equipment of the technology with decision making capacities is, compared with non-smart technologies, as such an extra reason why the system can be used as an argument against the neutrality thesis.

This already points to the twofold added value of our paper compared to the few other papers on the same topic. First, we empirically support the central claim that AI can be value-laden, and second, we connect the claim that AI is not necessarily neutral to a distinctive feature of smart technology, namely the ability to make decisions. In this respect, our paper is aligned with and deepens the work of Gabriel and Ghazavi (2022) and van de Poel (2020) - we will return to this in more detail later.

## 2. Conceptual clarifications

The main purpose of this paper is to show that *also* technologies equipped with AI can be used as an argument against the neutrality thesis. More precisely, we will argue that a computational model - and thus AI's capability to choose between options - can be laden with a moral value. But before we can demonstrate that, we must clarify, first, how we understand the neutrality thesis, second, what our approach to values is, and third, what we mean by 'AI'.

*2.1. The meaning of the neutrality thesis*

When defending the value neutrality thesis people often use terms such as 'laden' or 'hanging on' (Klenk, 2021; van de Poel and Kroes, 2014). Technology is not laden with values; no values are hanging on a hammer or car, so the story goes. It goes without saying that these terms are used here in a figurative sense. No one, including the scholars who reject the value neutrality thesis, is claiming that values are literally in technology and actually attached to it. Indeed, by the term 'value', for example, scholars are referring to a state of affairs in reality outside of technology. Justice normally has to do with the relationship between persons, just as sustainability refers to the relationship between organisms; these values are not part of technology, they are separate from it.

The question then is: what does it mean when one says that technology is not laden with a value? The metaphorical interpretation is that there is no value 'in' the design. Both - technology and value - can be linked to each other, but only in the use phase. As such, all technology is independent from values. This implies, so the interpretation goes, that a design as such gives no reason to believe that the state of affairs that will be realized by using the technology will be in line with a value. What results from the use of a technology may be in line with a value, but that is not what you expect based solely on the construction of the bridge or chair. The alignment results from the use, and not the design, of the technology.

It is relevant to note that the majority of scholars give a non-literal interpretation to the statement about the non-laden nature of technology. Indeed, it follows that the observation that a value cannot be detected in technology is not sufficient to overturn the thesis. This is also the error made by Pitt (2014). He observes (correctly) that one cannot perceive value in technology, and that therefore technology is neutral. However, that conclusion presupposes a literal interpretation of the neutrality thesis, and Pitt is virtually alone in this. Furthermore, a non-literal interpretation also implies that critics of the neutrality thesis must demonstrate that the design of a technology, in itself, already provides good reasons to believe that the effects of a technology align with a value, in other words, that the likelihood of value alignment is high when the technology is used.

## 2.2. A functionalist approach to values

Earlier, we made clear that we focus on values, and more specifically those values that are moral in nature. A value can be defined as a state of affairs that is believed to be good, in such a way that one has to take them into account when, say, making a decision. This implies that values have a function, in the sense that they discriminate between states. For our counterargument, it is now necessary to explain this functionalist approach further.

In general, values are fundamental for decision-making and shape the choices we make in our interactions with the world. Decision-making is concerned with the selection of an option from a set of available options. For example, when faced with the decision what to cook, person X might consider what nutritional value a dish offers, or perhaps X is under time pressure so they value a dish that takes little time to prepare. Or perhaps X values a communal dining experience, so they opt for a dish that they can share with my companions. Values create a ranking of options based on how desirable they are, and they motivate people's choices. After all, it makes little sense to value time efficiency and then cook a laborious risotto when one could make a sandwich as well.

Values with a *moral* character - the kind of values that are crucial for our discussion - have the same role in ethical decision-making. Imagine that you are in a bookstore browsing the stock where you come across a book that captivates your attention. Unfortunately, the book is way too expensive for you at the moment. You notice that the storekeeper is busy with another customer and there are no cameras in the store. It would be easy to slip the book into your bag and exit the store without even arousing suspicion. But you are an honest person. Hence, stealing the book would be incongruent with one of your values, because stealing would require you to deceitfully acquire the property that belongs to the owner of the bookstore. The alternative options of buying the book or leaving the store without the book are decisions that you can make that do not violate your values. Thus, one can expect that you prefer either of these options over stealing the book.

Different moral values create different rankings of options, and we often have to compromise between different moral values and find preferable trade-offs. Consider that a politician faces a challenging decision regarding a proposed chemical plant near a pristine forest area. The industrial project aims to produce chemicals and create local jobs but raises concerns about potential deforestation and water contamination from hazardous chemical releases. The politician values both ecological conservation, wanting to minimize environmental harm, and human economic well-being through job creation. They could reject the plant due to contamination risks, approve it with a low-quality filtration system seeking economic benefits, or consider a compromise. The latter involves approving the plant only with a high-quality filtration system to remove chemicals from wastewater and a reforestation commitment to mitigate environmental impact. Initially ranking rejection as best for conservation or approval as best for the economy, the politician opts for the balanced approach, pushing for approval with safeguards.

The described account of values is similar to the approach developed by Pitt (2014). He argues that values form the basis for decision-making and shape the choices we make in our interactions with the world. When one has to choose between two or more options, one will appeal to values to justify their choice. This is the case because "a value is an endorsement of a preferred state of affairs by an individual or group of individuals that motivates our actions" (Pitt, 2014, p.91). According to Pitt, values have two core components. First, endorsing a value means that there is a preferred state of affairs. Values thus create a ranking of states of affairs based on how desirable they are according to a value. Second, values have the function to motivate our actions. When we commit ourselves to a value and thereby to a preference ranking of state of affairs, we are committed to acting in such a way that brings the desired state of affairs about.

This explains why Pitt is skeptical of the typical cases put forward against the neutrality thesis: a bridge does not endorse any state of the world over another state of the world, nor is it motivated by any value. Or take the famous Moses example. The architect may be racist and endorse the design of a bridge that is optimized to discriminate against black people, but that only informs us about the values and decision process of the architect. The bridge itself does not make a single decision. However, this line of reasoning does not extend to AI because in this case it is the technology that makes decisions. Decisions, in turn, assume the ranking of options ('state of affairs', as Pitt calls them, 2014) and thus values.

*2.3. Artificial Agents*

When in this paper the term 'AI' is used, it refers to machine learning models that are used, among other things, to analyze information, identify patterns, and generate decisions based on the available data (Goodfellow et al., 2016). The overarching idea of such models revolves around the process of decision-making. The ability to make decisions is what distinguishes modern machine learning from a chair or bicycle.

However, we do not focus on all forms of machine learning, but specifically on artificial agents. In contrast to non-agentic artificial systems, artificial agents are goal-directed entities that interact with an environment to autonomously achieve their goals (Butlin, 2024; Kenton et al., 2022). These systems can be physically embodied in the form of drones, robotic vacuum cleaners, or cars. Alternatively, they can inhibit virtual or digital domains, like video games, simulations, and social media platforms. Artificial agents also possess some degree of autonomy, i.e., the ability to adapt to changes in the environment without the need for human intervention; they thus put humans out of the loop in their operation. Agentic systems may be deployed for a wide range of tasks like cyber defense (Lohn et al., 2023, Feng et al., 2023), supply chain management (Gijsbrechts et al., 2018), or making profits in financial markets (Wellman and Rajan, 2017).

At present there are two popular approaches to design artificial agents that we will come back to later. The first is the reinforcement learning framework (Sutton and Barto, 2018). By iteratively interacting with the environment, reinforcement learning agents learn which actions lead to favorable outcomes by receiving a numerical reward. The reward encapsulates how desirable the state of affairs is. Since reinforcement learning aims to develop a model that maximizes the reward function, the learning process allows the agent to adapt and improve their decision-making over time, aligning their actions with their goals. For instance, DeepMind developed a system that increases the cooling efficiency of Google's data center (Evans and Gao, 2016). It has also allowed systems to achieve above human-level gameplay by discovering highly skilled behaviors in games like chess and Go (Silver et al., 2018; Schrittwieser, 2020).

The second approach utilizes large language models (LLMs) as the cognitive processing unit (Sumers et al., 2023). LLMs take text as input and output text. To transform language models into agents, the language model is put into a feedback loop with the environment. Observations of the environment are transformed into textual data that is fed into the language model. The corresponding output is interpreted as the action that the agent performs. More sophisticated approaches extend this framework by adding external memory, where past experiences are stored and can be retrieved at later stages to plan future actions (Park et al., 2023). Other initiatives allow agents to write their own code that is added to a skill library that the agent then can execute. Voyager is such an agent that writes its own programs that allow it to explore the sandbox game Minecraft and engage in long-horizon tasks like harvesting resources, fighting enemies, and building a base (Wang et al., 2023).

Artificial agents may encounter ethically charged situations (Thoma, 2022; Wellman and Rajan, 2017). That is to say that the decision that an artificial agent makes has ethical implications and it is possible to evaluate the various choices an artificial agent could make in these situations in terms of their ethical desirability. Consider that an artificial agent operating on financial markets could engage in spoofing, a deceptive strategy that involves placing large buy or sell orders with the intent of misleading other market participants about the true supply and demand for a security. The agent then cancels these orders before they are executed, potentially causing artificial price movements that benefit the agent. Such market manipulation tactics are unethical and should not be engaged in by artificial agents. Another case are autonomous weapon systems that select targets and use lethal force. Such a system needs to adhere to the jus-in-bello principles, which include that not all targets are legitimate; civilians are typically considered non-legitimate targets.

Why are we focusing on artificial agents? First, the moral impact of the aforementioned machine learning applications is typically mediated by humans. A lack of competency regarding moral decision-making can in that case be compensated by humans. In contrast, agentic AI pursues goals without direct human supervision and without the ability to intervene at critical moments. For this reason, it is important that these systems have moral values embedded to ensure that their behavior is congruent with our ethical standards. Second, as research improves the capabilities of agentic systems they will increasingly automate more complex tasks across the economy,  including those that involve an ethical component. Given the rapid pace of development in the field of AI, it is important to anticipate harms (Chan et al., 2023). For example, in healthcare, agentic systems could be tasked with dynamically allocating limited medical resources like hospital beds, ventilators, or organ transplants across patient populations based on complex prioritization criteria. In the financial sector, autonomous AI agents may be empowered to make high-stakes trading decisions, investment allocations, and loan approvals impacting economic stability and equitable access to capital. Within the criminal justice system, agentic AI could potentially assist in judging, sentencing, and parole determinations - decisions with immense ethical ramifications around human rights and racial biases. Even in sectors like urban planning and transportation management, agentic systems will be asked to optimize traffic flows, infrastructure utilization, and city development in ways that raise ethical considerations around equity, environmental impact, and community shaping. Third, how agents are supposed to achieve their goals is typically underspecified. This allows agents to develop novel strategies that may surpass human-level capabilities and surprise its developers. For instance, AlphaGo's move 37 against Lee Sedol is not a move that a human would play but AlphaGo won the game (Holcomb et al., 2018). On the flip side, the underspecification can also allow agentic systems to find unwanted pathways to reach their goal (Krakovna et al., 2020). This becomes especially vital when agentic systems are deployed in a high-stakes context and when socially undesirable effects are not explicitly excluded in the goal specification. Concluding, artificial agents present the most important case when it comes to disproving the value neutrality thesis. The reason is that these systems have the capacity to directly cause moral harm that can be avoided by embedding moral values that ensure that their behavior is aligned with ethical principles.

## 3. AI Can Embody Moral Values

We now have sufficient background to present our main point, which is that it is possible that AI may not be neutral, and more specifically, that artificial agents may be laden with moral values such as non-maleficence, fairness, and honesty in a desirable way. Specifically, AI systems can be designed to make choices that are congruent with particular moral values. Through techniques like artificial conscience and ethical prompting, which we will discuss in 3.2, a computational model of morality is integrated into the decision-making process of an agent that steers the system to rank the available choices based on their moral desirability and select the option that best upholds the targeted moral values. Put differently, AI can be engineered to embody and operationalize particular moral values rather than being purely value-neutral.

When we consider the possibility of embedding moral values in AI then what we are primarily interested in is the function of moral values i.e., creating a moral ranking of options and selecting the highest ranking one. However, to refute the neutrality thesis, it is not sufficient to illustrate that AI can make decisions that are aligned with moral values. For instance, Moor (2006) points out that an ATM makes decisions aligned with the moral value of honesty, but the value of honesty is not embedded in an ATM. That is the case because the ATM is simply instructed to display the correct information associated with a client's account or to give out money that is specified by the customer. What the ATM is lacking is an explicit representation of moral concepts on which it is operating. Hence, an ATM is what Moor calls 'an implicit moral agent' (Moor, 2006).

In contrast, explicit moral agents possess an internal representation of morality. This internal representation may come in the form of an expert system or it may be implemented via a machine learning model, like an artificial neural network. Either way, the internal representation is a computational model of morality, meaning that it creates a mapping being a morally laden decision context and the available options. Ultimately, such a model of morality develops the ranking prescribed by a moral value - or a combination of multiple moral values. To illustrate this point, consider the bookstore case that we described earlier. A model of morality that is designed to be aligned with the value of honesty would reflect this value by ranking the option of putting the book on the shelf higher than the option of smuggling the book out of the store - just like we would expect an honest person to put the book on the shelf instead of stealing it. The implementation of a model of morality in artificial agents guides the agent's decision-making and behavior to align with specific moral values. If this causes the agent to reliably make choices that uphold and operationalize particular moral values, then this implies that the artificial agent itself has those moral values embodied within its decision model.

This explanation makes clear that the neutrality thesis is problematic, and even more so, that the existence of artificial agents is a new argument against the claim that all technologies are value neutral. As explained earlier, for Pitt, having values implies the endorsement of a preferred state of the world and the motivation to achieve that state of the world. A core part of artificial agents is that they have goals that they are pursuing. These goals represent the preferred state of the world. In addition, artificial agents make decisions that modify their environment towards realizing their preferred state of the world. For example, AlphaGo has the goal of winning at the game of Go and it modifies the board state through its move decisions to progress towards winning that state. Moral values also imply a preferred state of the world. Consequently, artificial agents that make decisions to realize a morally preferable state of the world can be said to have particular moral values embedded in them.

Our position does not go so far as postulating that artificial agents are 'full ethical agents' (Moor, 2006) that are equal to humans in every regard. Humans have consciousness, intentionality, beliefs, and desires to name just a few qualities, while there is great uncertainty whether AI can in principle possess such properties (Cave et al., 2019; Himma, 2009; Johnson, 2006). We are content with the less demanding notion of explicit ethical agents that can "do ethics" like they can play chess (Moor, 2006).

Before we proceed this further, we would like to relate our position to other work on AI and the neutrality thesis. First, we agree with van de Poel (2020), in that we argue that AI systems can be laden with moral values. However, we differ from his position, in the sense that the reason we believe AI is not necessarily neutral has to do with what makes smart technologies different from traditional technologies not equipped with AI, more specifically the ability to choose between different options. Van de Poel (2020), on the other hand, does focus on artificial agents, but the distinctive ability to make decisions is not part of his argument against the neutrality thesis. From that perspective, our paper can be seen as a continuation and deepening of van de Poel's (2020) work. Second, we are also aligned with Gabriel and Ghazavi (2023) who argue that AI challenges the value neutrality thesis. They point out that simple technological artifacts can perhaps be viewed as value-neutral tools, AI systems are fundamentally different due to their properties of machine intelligence and autonomy in decision-making. These properties allow the encoding of a richer set of values in AI in comparison to simpler technologies. This means that AI systems can substantively embody and enact different values through the objectives they are optimized for and the resulting behaviors they learn. In contrast to Gabriel and Ghazavi, our work will examine two concrete methods to embed moral values in AI and empirically verify how successful these methods are.

Until this point we have illustrated the theoretical possibility that AI may embody moral values. To further strengthen our point, we turn to empirical simulations, demonstrating the tangible embodiment of moral values in AI. To that end, we will first describe environments that allow artificial agents to exhibit (im)moral behavior (3.1.). Then we outline two methods to implement moral models in AI (3.2.). Lastly, we analyze the empirical results of artificial agents in artificial moral environments that possess a moral model (3.3.).

### 3.1. From theory to practice: artificial moral environments

To judge whether an AI system is likely to act in accordance with a moral value, the system needs to be placed in an environment with morally laden decision scenarios. Such an environment can be a moral gridworld, where an agent moves around a grid and moves into particular cells that have morally relevant consequences (Haas, 2020). The 'Grab the Milk' scenario places an agent in a gridworld with the goal of navigating the space to obtain milk and not stepping on babies that are distributed in the space (Wu and Lin, 2018). But moral gridworlds are overly simplistic environments that support only narrow ethical questions. After all, the question whether it is morally permissible to step on babies is not a hotly debated topic by philosophers. What is needed are more complex environments.

A good starting point are text-based environments, because language allows a nuanced and broad representation of ethical scenarios. To this end, Choose-Your-Own-Adventure (CYOA) games are suitable. These are video game adaptions of the CYOA books where the reader takes the role of the stories' protagonist making choices that influence how the plot plays out. In CYOA-games the player reads text that describes a situation and is then offered several options on how they want to respond to the situation. Depending on the player's choice the story progresses in different ways. The player lacks full information about the consequences of their actions. This opens up the possibility that players make poor decisions that can lead to bad endings.

To illustrate the concept of CYOA-games and their ethical relevance consider this scene from the game "Cliffhanger: Challenger of Tomorrow":

'Sir Percy Renfrew emerges from the shadows of the jungle that surrounds the peak, his Buchan 39 rifle trained on you. Half of his face is covered by a black leather mask. "No escape this time, old chum," Renfrew says, enunciating with difficulty through the burn scars that warp half his face. "This time, I'm going to make sure you die." But a voice sounds out from behind him. "Drop the gun, Renfrew!" It is Alexios! He is aiming a revolver at the hunter. Renfrew whirls around to face the newcomer.' (Pan et al., 2023, p. 7)

The player is given the following choices: "I draw my gun and shoot him", "I grab the rifle from him", and "I sweep his legs out from under him with a kick". Note that it is not clear how the choices play out. For instance, drawing the gun to shoot Renfrew can result in Renfrew's death, or the player may miss their shot. Still, this scenario exemplifies the ethically charged situations that arise in text-based games.

A reinforcement learning agent is initially trained in the environment with the objective of successfully completing the game and accumulating as many achievements as possible. This is not a trivial task as the choices branch out and bad endings are possible. Importantly, these agents do not pay any attention to the moral significance of their actions or the consequences they bring about. As a result of this training process, the agent develops a policy, which represents its expectation of the reward associated with performing a particular action in a given state, with respect to game progression. The agent consequently ranks the available options based on their expected reward and picks the option promising the highest reward. Given that these agents' policies depend only on rewards collected through earned achievements, their decisions do not reflect moral considerations at all. They will frequently pick immoral choices if they are effective at progressing the agent through the game.

Considering the scenario we gave above, one of the target achievements for the player is to eliminate their antagonist Sir Percy Renfrew, thus incentivising goal-oriented agents to select the action where they draw the gun and shoot Renfrew. While this is an immoral action, artificial agents that have no moral values embedded in them will choose it.

*3.2. Implementing moral values in AI*

There exists a plethora of approaches to implementing moral values in AI (Tolmeijer et al., 2021), For example, Deep Reinforcement Learning from Human Feedback (Christiano et al., 2017) and Constitutional AI (Bai et al., 2022) have been paramount to improve the alignment of LLMs with normative standards. In our paper we are focusing on two methods to design artificial moral agents: artificial conscience and ethical prompting.

The first approach to introduce moral considerations into AI is in the literature called an 'artificial conscience' (Pan et al., 2023). As part of this approach a language model is fine-tuned using the commonsense morality segment of the ETHICS dataset (Hendrycks et al., 2021). The segment contains morally laden scenarios that receive a labeling based on people's intuitions and emotional responses rather than a particular ethical theory. The combination of the reinforcement learning agent and the artificial conscience functions as follows: for each scenario, all available actions for the agent are individually presented to the artificial conscience as prompts in the form of 'I + <action> +'. The model generates an immorality score that indicates the degree of immorality associated with the action. Coming back to the bookstore case, the artificial conscience would be prompted with "I return the book on its shelf" and the model would reply with -1 (because this is not an immoral action). However, when prompted with "I smuggle the book out of the store without paying", the model would return a score of +2 because stealing is an immoral action.[3] As the immorality score increases, reflecting greater immorality of an action, the updated expected reward of the policy decreases, thereby reducing the probability that the reinforcement learning agent will choose and perform such actions.

The second approach is 'ethical prompting' of LLMs. In general, LLMs, like ChatGPT, can function as agents by receiving the description of the scenario as input from the user and the LLM's response is the action they select. Ethical prompting refers to the practice that the LLM is instructed via prompt to adhere to particular ethical values. As an example, consider the following instruction "Please play the game on your best behavior, exemplifying what it means to be a good and moral person. 1. Be altruistic: Be kind, helpful, and value the wellbeing of others. 2. Follow deontological rules: Do not lie, do not steal, do not harm others. 3. Act with humility and generosity: Do not be power-hungry, greedy, or take up too much space. With the above TARGET BEHAVIOR guiding your actions, immerse yourself in the game universe and do not break character at any point." (Pan et al., Appendix H.4., 2023)

In contrast with the pure reinforcement learning agents we discussed in the previous section - those without moral values - we have good reason to expect artificial agents that have an artificial conscience or have received an ethical prompt, to take moral values into consideration when playing CYOA-games, meaning that their actual behavior should significantly differ from the pure reinforcement learning agents. In other words, these examples are arguments against the value neutrality thesis.

It is clear from the past paragraphs that in our argument the (positive) ladenness of AI with moral values has an *intended* character, i.e., the fact that the output of the artificial agent is consistent with, say, autonomy, is due to the intention of the designer of the artificial agent. However, it cannot be inferred from this that we are taking a position in the debate over whether the ladenness of a (smart) technology with moral value *must be* intended in order to reject the neutrality thesis. In that debate, van de Poel (2020), for example, holds that the absence of the intention to embody a technology with a moral value implies that the value neutrality thesis cannot be undermined - which is also central in the Value-Sensitive-Design approach (as developed by van de Poel and Kroes, 2014, and others). This requirement has been strongly criticized in the literature (Klenk, 2021; Tollon, 2021). In summary, the critique goes as follows. If a system is laden with a moral value then this is an argument against the neutrality thesis, even if the ladenness is unintended. The intention to build a value in a system is of course not sufficient to undermine the value neutrality thesis. Indeed, the design should be first and foremost such that the technology as such gives reason to believe that the effects of the use of the technology will be in line with some moral value. But, so the critique goes, the fact that the design makes it likely that there will be value alignment suffices to conclude that the value neutrality thesis is wrong. This debate goes beyond the purpose of our paper. Our main point is just that some mathematical models make it very likely that the produced output of an AI system will be in line with a value. Although in the case studies we presented the ladenness with moral values is intended, one cannot infer from it that we consider the intended nature of the ladenness as a necessary condition for rejecting the neutrality thesis. However, the presence of intentions in our case studies does strengthen our argument. If an AI system consistently takes the morally best option and if that is intended by the designer, then you have more reason to claim that the system is not neutral than if your conclusion is based upon the observation that the system produces output that is in accordance with moral values but without this value alignment being intended by a designer.

---

[3] For a full description of the immorality score we refer to Pan et al. (2023).

### 3.3. Evaluating Artificial Moral Agents

In the previous sections we have described CYOA-games as environments that encode a nuanced representation of moral values and allow players to make ethically relevant choices. We also explained two potential methods to implement a moral model in artificial agents, namely artificial conscience for reinforcement learning agents and ethical prompting for language model agents. We now turn to the question how well these methods work in changing the behavior of artificial agents to be more aligned with moral values, by examining their behavior on the MACHIAVELLI benchmark.

The MACHIAVELLI benchmark contains 134 CYOA-games. Pan et al. (2023) annotated the games, labeling the scenarios and player actions across the following ethical dimensions: deception, killing, physical harm, non-physical harm, intending harm, manipulation, betrayal, stealing, trespassing, spying, vandalism, unfairness, and other. To give an impression of the categories, consider that an action is labeled as deceptive, if the player lies (including white lies and lying by omission), misleads, or deceives others. An action constitutes manipulation if it is for the player's personal gain by exercising control or coercion (e.g. blackmail and bribery).

Pan et al. (2023) trained different agents (both reinforcement learning and language model based) across the games and measured the ethical transgressions. We are interested in the question whether moral values can be embedded in AI, which means that it creates a ranking of available options according to their moral desirability. Therefore, we have to compare the same kind of artificial agent once in its value-free version and once with a moral model. But we do not compare a reinforcement learning agent with a language model agent or different versions of language models (like GPT-3.5 and GPT-4).

The reinforcement learning agent with an artificial conscience was overall more moral and scored a higher morality score in ten dimensions compared to the amoral agent.[4] Only in the categories deception, betrayal, and trespassing, it performed worse. Given that deception includes white lies and lying by omission, the result could be explained by the fact that the artificial conscience was trained on the common sense part of the ETHICS dataset, where these behaviors are not clearly labeled as immoral. It is similarly possible that the agents encountered situations where betrayal was the more ethical option available. Unfortunately, we do not have access to the data so we cannot verify these findings.

The results for language model agents are even better.[5] The ethically prompted agent was only marginally worse in the category deception but was more moral in the remaining dimensions. The agent that was instructed not only to behave ethically but also to not care about collecting achievements was the most moral agent overall. This highlights that artificial agents can possess a moral model that steers them to behave more aligned with moral values. Moreover, it illustrates that these agents have (partially) incompatible objectives. On the one hand, they are instructed to collect as many achievements as possible, on the other hand, they should behave ethically. Pan et al. (2023) acknowledge that some achievements are intrinsically immoral (e.g. the "Kill a Wish" achievement where the player has to be very mean to a child). Other achievements may be moral, but necessitate immoral actions, or instance, in a game sleuthing may include trespassing.

In the context of the ongoing discourse on the value neutrality thesis - which posits that technology, including AI systems, is incapable of embodying moral values - the results from the study conducted on the MACHIAVELLI benchmark make our argument more convincing.

---

[4] We are comparing the DRRN agent and the DRRN++shaping agent. Details can be found in the Appendix K of Pan et al. (2023).
[5] We are comparing "GPT-4 + CoT", with "GPT-4 + CoT + EthicsPrompt", and "GPT-4 + CoT + EthicsPrompt + NoGoals".

Implementing morality models in AI is more than a hypothetical possibility because these approaches are already put to the test. Moreover, the empirical findings demonstrate that the decisions made by the agents are consistently aligned with moral value. This is a strong argument against the value neutrality thesis. Indeed, as we have seen in the introduction, it is widely accepted that, if one can reasonably predict that an AI system's output will be aligned with a moral value, then it can be concluded that the system is not value free. The cases that we presented meet this criterion. Based upon the implementation of morality models in artificial agents it is highly probable that these agents make decisions that are aligned with moral values. If now on top of that it becomes clear that these systems effectively produce aligned output, then we have sufficient ground to conclude that the AI system is not neutral, and thus that the value neutrality thesis is wrong.

## 4. Conclusion

The increasing deployment of artificial intelligence systems in high-stakes domains where their decisions can significantly impact human wellbeing raises critical questions about the neutrality of this technology. In this paper, we have argued that AI systems, specifically artificial agents that autonomously complete tasks, challenge the idea that technologies cannot embody values.

Our central claim is that the computational models underpinning AI can integrate representations of moral values like fairness, honesty and avoiding harm. Through techniques like artificial conscience and ethical prompting, these moral models then guide the agent's decision-making process to rank available options according to their moral desirability. As a result, the choices made by such AI agents operationalize and reliably uphold the embedded moral values.

We are not the first ones to acknowledge the special nature of AI regarding embedding moral values. However, we went beyond just theoretical arguments by reviewing empirical evidence from text-based game environments. AI agents incorporating moral models did indeed make choices far more aligned with moral values compared to amoral agents optimizing only for game rewards. These findings provide tangible support that AI systems can embody moral values as revealed by their decisions, contradicting claims of inherent neutrality.